\documentclass[10pt,twocolumn,letterpaper]{article}

\usepackage{wacv}

\usepackage{hhline}
\usepackage{xcolor,colortbl}
\usepackage{graphicx}
\usepackage{amsmath}
\usepackage{amssymb}
\usepackage{booktabs}
\usepackage{multirow}
\usepackage{makecell}
\usepackage[numbers]{natbib}

\definecolor{lightergray}{cmyk}{0,0.0,0.0,0.14}
\usepackage{array}
\usepackage{pifont}
\usepackage{wrapfig}
\usepackage[pagebackref,breaklinks,colorlinks]{hyperref}
\newcommand{\cmark}{\ding{51}}
\newcommand{\xmark}{\ding{55}}
\newcolumntype{?}{!{\vrule width 2pt}}

\usepackage[capitalize]{cleveref}
\crefname{section}{Sec.}{Secs.}
\Crefname{section}{Section}{Sections}
\Crefname{table}{Table}{Tables}
\crefname{table}{Tab.}{Tabs.}

\def\wacvPaperID{852} 
\def\confName{WACV}
\def\confYear{2025}

\makeatletter
\renewcommand{\paragraph}{%
  \@startsection{paragraph}{4}%
  {\z@}{0.5ex \@plus 0.5ex \@minus .2ex}{-1em}%
  {\normalfont\normalsize\bfseries}%
}
\makeatother

\begin{document}

\title{LowFormer: Hardware Efficient Design for Convolutional Transformer Backbones }

\author{Moritz Nottebaum\textsuperscript{1}\\
{\tt\small nottebaum.moritz@spes.uniud.it}
\and Matteo Dunnhofer\textsuperscript{1}\\
{\tt\small matteo.dunnhofer@uniud.it}
\and Christian Micheloni\textsuperscript{1}\\
{\tt\small christian.micheloni@uniud.it}
\\
\textsuperscript{1}University of Udine, Italy\\
}

\maketitle

\begin{abstract}
Research in efficient vision backbones is evolving into models that are a mixture of convolutions and transformer blocks. A smart combination of both, architecture-wise and component-wise is mandatory to excel in the speed-accuracy trade-off. 
Most publications focus on maximizing accuracy and utilize MACs (multiply accumulate operations) as an efficiency metric. The latter however often do not measure accurately how fast a model actually is due to factors like  memory access cost and degree of parallelism.
We analyzed common modules and architectural design choices for backbones not in terms of MACs, but rather in actual throughput and latency, as the combination of the latter two is a better representation of the efficiency of models in real applications.
We applied the conclusions taken from that analysis to create a recipe for increasing hardware-efficiency in macro design.
 Additionally we introduce a simple slimmed-down version of Multi-Head Self-Attention, that aligns with our analysis. 
 We combine both macro and micro design to create a new family of hardware-efficient backbone networks called LowFormer. 
   LowFormer achieves a remarkable speedup in terms of throughput and latency, while achieving similar or better accuracy than current state-of-the-art efficient backbones. In order to prove the generalizability of our hardware-efficient design, we evaluate our method on GPU, mobile GPU and ARM CPU.
We further show that the downstream tasks object detection and semantic segmentation profit from our hardware-efficient architecture.
Code and models are available at \url{https://github.com/altair199797/LowFormer}.

\end{abstract}

\section{Introduction}
\label{sec:intro}

\begin{figure}[hbt!] 
  \includegraphics[width=\linewidth]{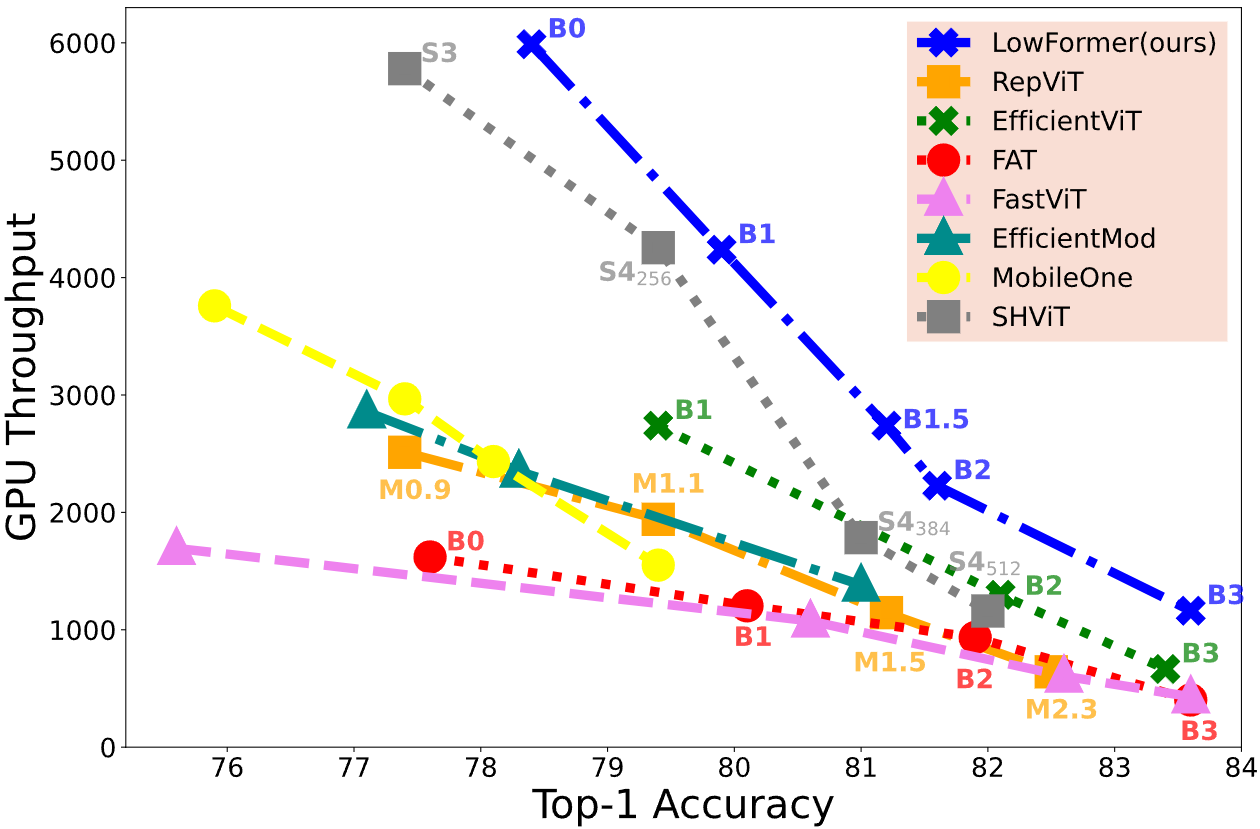}
  
  \caption{Comparison of GPU throughput and top-1 accuracy of recent image classification architectures and LowFormer. The markers refer to different complexity classes of corresponding architecture families. LowFormer consistently achieves a higher throughput than models with similar accuracy. }
  \label{fig:throughputcompfig}

\vspace{-0.3cm}
\end{figure}

Recent research in efficient vision backbone networks has focused on combining attention mechanisms with convolutional layers. The mixture of local information extraction (by convolutions) and global reasoning (by attention) has proven to be superior to homogeneous models \cite{efficientvit, efficientvitbad,mobilevit,shvit}. 
Besting the speed-accuracy trade-off has been the primary goal of research in that domain, in order to subsequently improve the efficiency of downstream tasks, like object detection, pose estimation and semantic segmentation \cite{battleofthebackbones,effbbobjectdetec}. Efficiency in backbone architectures is also particularly important to improve the applicability of downstream tasks on mobile and edge devices \cite{mobileone}. 

To measure the computational efficiency of deep learning models and compare them, it is common to count the amount of MACs (multiply and accumulate operations) \cite{efficientnet}.
The number of MACs of a model also strongly correlates with its accuracy.
For example, when increasing them by width, depth or resolution scaling, the accuracy can be improved \cite{efficientnet}.
However MACs are not created equal and ignore factors like memory access cost and the degree of parallelism \cite{hardwareffficientdesign,efficiencymisnomer, mobileone,efficientvitbad}, which can have a substantial effect on execution time of a model.

Our goal is to improve throughput and latency for efficient backbone architectures while increasing accuracy by investigating when MACs are executed most efficiently. We subsequently deduct recipes for efficient architectural design, that follow these findings.
For our investigation we carry out several experiments in \Cref{sec:speedexperiments} that contrast execution time and number of MACs of macro design decisions. 

We apply the findings to create a new family of backbone networks called LowFormer.
The exploitation of efficient MAC execution due to our macro design
makes LowFormer faster and more accurate than previously proposed models (see \Cref{fig:throughputcompfig}).
Following the results of our investigation we complement our architecture with 
 a simple and lightweight adaption of the traditional attention \cite{attentionisallyouneed}, that downsamples and upsamples the feature maps around the Scaled Dot-Product Attention (SDA). As a result the SDA is operating on a lower input resolution, hence the name \textbf{Low}Former. 

We confirm the generalizability of the hardware efficiency  of LowFormer by measuring GPU throughput, GPU latency, mobile GPU latency and ARM CPU latency.
We back up our design decisions by an extensive ablation study (see \Cref{subsec:ablation}). 
Our proposed architecture has a simple micro and macro design and allows us to scale it from low complexity(LowFormer-B0) to higher complexity(LowFormer-B3). In total we feature five models (B0,B1,B1.5,B2,B3) and our top-1 accuracy on ImageNet-1K \cite{imagenet}  ranges from 78.4\% to 83.64\%.
For example compared to MobileOne-S2 \cite{mobileone}, LowFormer-B0 has 2$\times$ the throughput and 15 \% less latency on GPU, while scoring 1\% better in top-1 accuracy.
Our model LowFormer-B3, featuring the highest complexity, almost has 3$\times$ the GPU throughput and 55\% of the GPU latency of FAT-B3 \cite{fat}.

We also integrate LowFormer into Semantic FPN \cite{semanticfpn} and RetinaNet \cite{retinanet} to improve the efficacy of semantic segmentation and object detection models. Within the first framework for example, our LowFormer-B1 backbone achieves 1.7\% better mIoU than FastViT-SA12 \cite{fastvit}, while having 3$\times$ the throughput and 30 \% less latency on GPU.
In summary, our contributions are as follows:
\begin{itemize}
     \vspace{-0.2cm}
    \item We present a new backbone architecture family with a hardware-efficient macro design and a new lightweight attention. Our models are faster in terms of throughput and latency compared to models with similar accuracy.\vspace{-0.2cm}
    \item 
    On several computing devices we carry out an exhaustive  speed analysis of convolutions in different configurations to contrast amount of MAC operations and measured execution time. 
\\
     
     \item We show that LowFormer generalizes well to semantic segmentation and object detection, as well as retains its speed-up on different hardware like Mobile GPU and ARM CPU.
\end{itemize}

\section{Related Work}
\label{sec:relatedwork}

\paragraph{Hardware-Efficient Model Design.}

Achieving the highest accuracy at all (computational) cost has long since ceased to be the only goal in deep learning \cite{efficientnet}. An ever-increasing share of research focuses to create the most efficient architecture and subsequently to achieve the best speed-accuracy trade-off \cite{edgevit, mobilevig, mobilenetv3}. Earlier approaches mainly equated speed with a minimal amount of MACs \cite{ghostnetv2,ghostnetv1,mobileformer}, while more recent research increasingly judges models by throughput or latency on for example desktop GPU and CPU \cite{shufflenetv2,shvit,efficientvit}, mobile GPU \cite{mobileone} or generally edge devices \cite{edgevit}.

Memory access cost and degree of parallelism have become important factors for efficient model design.
Methods like EfficientViT \cite{efficientvitbad} remark that the Multi-Head Self-Attention (MHSA) induces higher memory access cost than the Feed-Forward Network (FFN) in the transformer block. Therefore they increase the ratio of FFN compared to MHSA in their architecture, with minimal accuracy loss. 
The authors of MobileOne \cite{mobileone} on the other side analyze the effect of activation functions and multi-branch architectures on mobile latency. 
ShuffleNetV2 \cite{shufflenetv2} and FasterNet \cite{fasternet} both point out that grouped convolutions are inefficiently executed on current hardware due to high memory access cost \cite{fasternet}.
We follow their insights, but instead of ungrouping a portion of the convolutions \cite{shufflenetv2} or introducing a new micro design \cite{fasternet}, we study how fusing depthwise and pointwise convolutions affects execution time.
We further analyze the effect of resolution on hardware-efficiency and apply the insights of both studies in our macro design.


\paragraph{Efficient Attention.}
There is a jungle of different kind of attention-like operations, that try to replace the traditional attention operation. Many remove its quadratic nature by variations of linear attention \cite{hydraattention,efficientvit, maxvit} based on \citeauthor{linformer}\cite{linformer}. 
However \citeauthor{poolformer}\cite{poolformer} showed that attention in itself is not as important as we thought and can even be replaced by a simple pooling operation. \citeauthor{efficientformer}\cite{efficientformer} took that idea further and used the efficient pooling operation for the first 3 stages  and the traditional attention for the last 2 stages \cite{efficientformer}. 
Others \cite{CvT,PyramidTransformer,multiscalevit} downsample the keys K and V before the attention operation, either with convolutions or pooling . 
 \citeauthor{inceptiontransformer}\cite{inceptiontransformer} on the other side also downsamples Q, therefore completely operating on a lower resolution. This is similar to how our attention approach works, however we use convolutions for downsampling instead of pooling and do not incorporate their inception token mixer, which separates the channels for multiple paths, one of them being the attention on a lower resolution. Additionally in contrast to us \citeauthor{inceptiontransformer}\cite{inceptiontransformer} apply their inceptionformer block in all stages. They also miss other optimizations which we describe in \Cref{subsec:microdesign}.
 
 Our adaptation of  MHSA is simple and is close to the traditional attention, setting aside overly complicated approaches to an already established concept. 
 To the best of our knowledge nobody yet proposed this adaptation to the original Multi-Head Self-Attention \cite{attentionisallyouneed}.

\section{Execution Time Analysis}
\label{sec:speedexperiments}

In the following we will investigate the hardware-efficiency of convolutions in different configurations. 
A convolution or a block of convolutions is for example more hardware-efficient than another, when it has more MAC operations, but is at least similar fast.
We will evaluate speed on desktop GPU and CPU. 

In \Cref{subsec:macrodesign} we describe how we apply the insights gained from the following experiments.

\subsection{Depthwise Convolutions}
\label{subsec:grouping}

When the focus of a model is efficiency and mobile-friendly design, depthwise convolutions are a prominent alternative to standard convolutions \cite{mobilenetv3, efficientnet}. Standard convolutions are ungrouped convolutions (groups=1), while depthwise convolutions are grouped convolutions that have as many groups as input channels.
While depthwise convolutions are efficient in terms of MACs, they can not translate latter completely to common hardware. Evaluating models by their MACs in architecture design automatically leads to inserting as many depthwise convolutions as possible, regardless of their effective speed-up.
In order to concretize the disconnection of execution time and MACs as an efficiency measure, we carry out an experiment using a simplified toy architecture. 
We examine the effect of using depthwise convolutions on GPU throughput, GPU latency and CPU latency (see  \Cref{tab:groupingexp}). We created three models with only depthwise convolutions (\#2,\#4,\#6)  and three with 
only standard ones (\#1,\#3,\#5). 
Even though model \#1 and \#2 in \Cref{tab:groupingexp} have a similar GPU throughput and Latency, \#1 (ungrouped) has 4$\times$ the amount of MACs. The same is true for model \#3 and \#4. Model \#5 (ungrouped) even has 6$\times$ the amount of MACs, while having a slightly higher throughput. Regarding GPU latency however, the models with ungrouped convolutions (\#1, \#3 and \#5) are slightly slower.

In summary, depthwise convolutions are not as hardware-efficient as standard convolutions. They do not give the expected speed-up which their low amount of MACs might suggest.



\begin{table}[t]
    \centering
    \resizebox{8cm}{!}{
    \begin{tabular}{c|c|c|c|c|c|c}
        \Xhline{1.5pt}
         \multirow{2}{*}{Model} & Channel & \multirow{2}{*}{Depthwise} & MACs &  GPU Throughput & \multicolumn{2}{c}{Latency }  \\ 
         & {[C$_0$, C$_1$, C$_2$, C$_3$, C$_4$]} & &  (M) & (images/s) & \multicolumn{1}{c}{CPU } & \multicolumn{1}{c}{GPU } \\
         
         \Xhline{1.0pt}
         \#1 & {[17, 34, 68, 136, 273]}  & \xmark & 887 &  \textbf{5263}  & \textbf{7.22} & 0.26 \\
         \#2 & {[30, 60, 120, 240, 480]} & \cmark \par & 207  & 5025  & 10.05 & \textbf{0.24 } \\
         \Xhline{1.0pt}
         \#3 & {[32, 65, 130, 260, 260]}  & \xmark & 2734  & 2617 & \textbf{15.36} & 0.51  \\
        \#4 & {[60,120,240,480,480]}  & \cmark \par & 750  & \textbf{2624}  &  20.20 & \textbf{0.45} \\
        
        \Xhline{1.0pt}
         \#5 & {[32, 96, 193, 387, 387]}  & \xmark & 5270  & \textbf{1886}  & \textbf{22.32} & 0.74 \\
         \#6 & {[60, 180, 360, 720, 720]} & \cmark \par & 809 & 1805  &  25.84 & \textbf{0.65} \\

        \Xhline{1.5pt}

    \end{tabular}
    }
    \caption{Comparison of throughput and latency of depthwise and standard convolutions in relation to amount of MACs. The six toy models (\#1-6) are divided into three groups by similar throughput. Each model has the same architecture and only differs in the number channels (see second column) and if its convolutions are depthwise or not (see third column). The table shows, that standard convolutions that feature many more MACs (4$\times$ to 6$\times$ the amount) can still be similarly fast as depthwise convolutions. Bold entries refer to the best value in each group. } 
    \label{tab:groupingexp}
    \vspace{-0.3cm}
\end{table}

\begin{figure}
\centering
MBConv with expansion factor 4

\vspace{0.1cm}

\includegraphics[width=\linewidth]{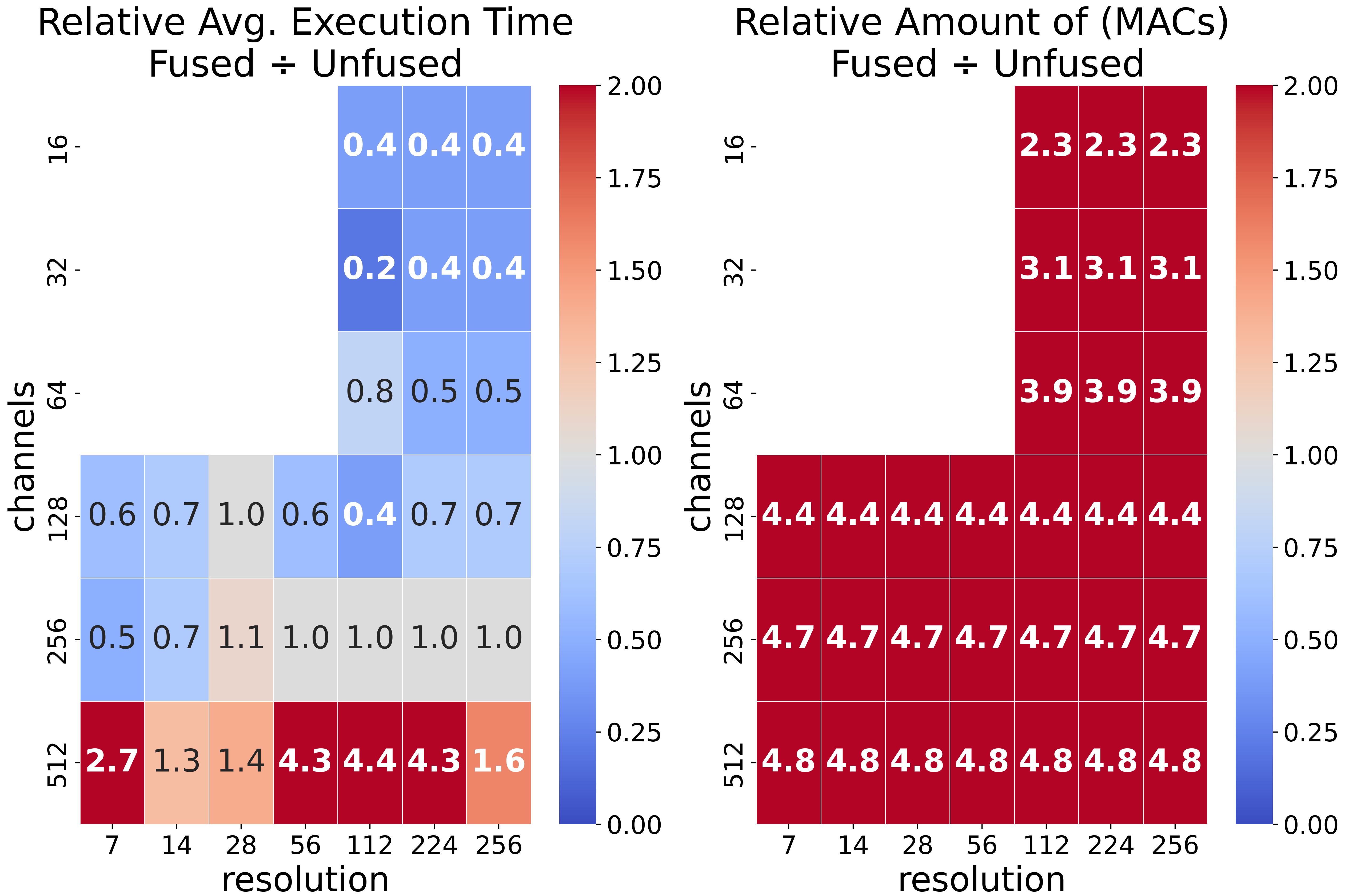}
\vspace{0.3cm}

MBConv with expansion factor 6

\vspace{0.1cm}

\includegraphics[width=\linewidth]{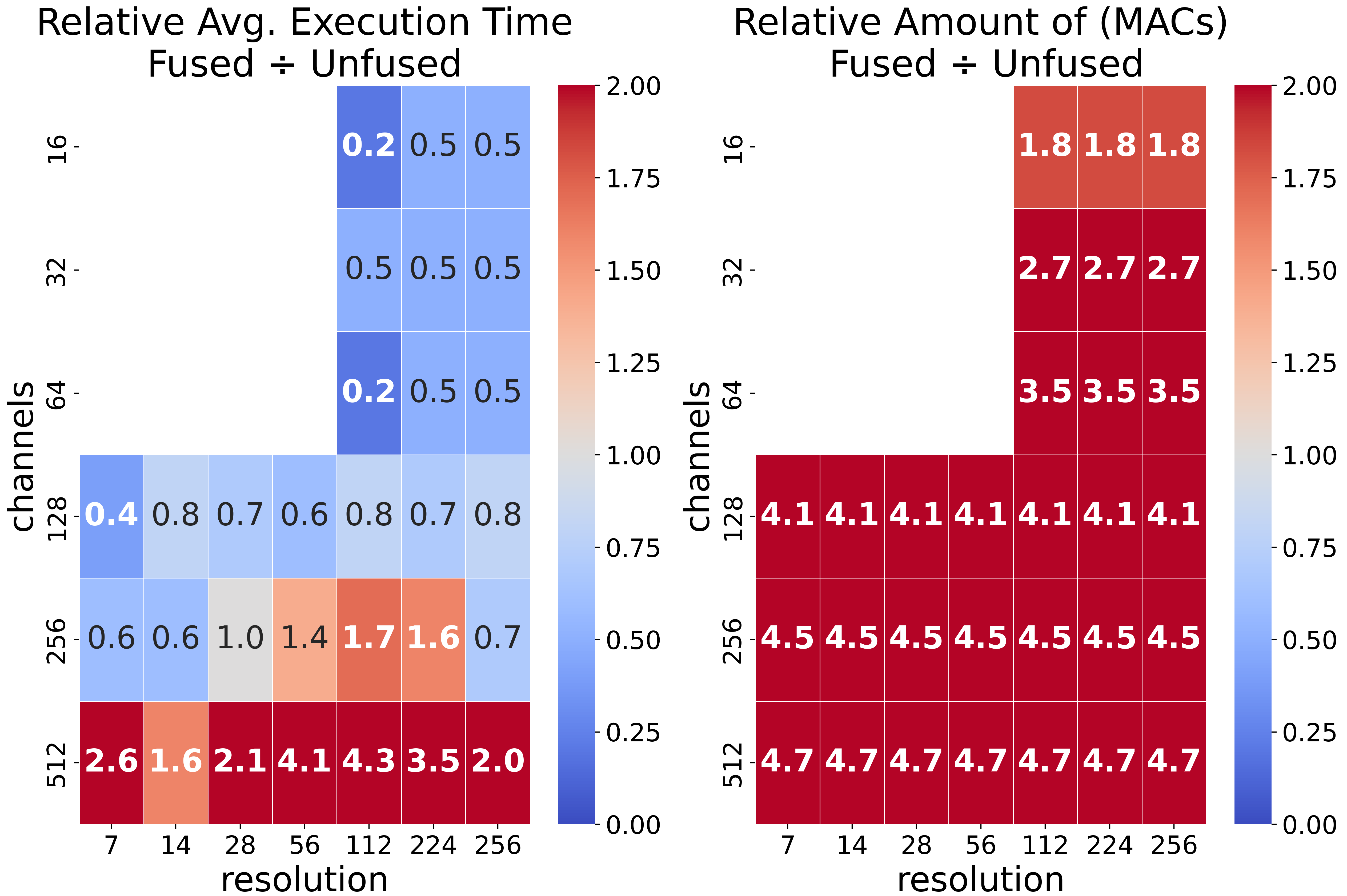}
  \caption{ The left figures depict average execution time on GPU  of the fused mobile inverted bottleneck (MBConv)
relative to the unfused one, while the right figures depict the relative amount of MACs.  The blue areas in the left figures correspond to configurations (number of channels and resolution) where fused MBConv is faster, while red corresponds to the opposite. For the right figures the red areas correspond to configurations where the fused MBConv has a higher amount of MACs. Bold and italic numbers refer to entries with a particularly unequal ratio. 
Even though the fused MBConv always has more MACs, it is faster for many configurations. }
  \label{fig:fusedexp}
\vspace{-0.3cm}
\end{figure}

\subsection{Fused vs. Unfused MBConv}
\label{subsec:fusedunfused}

The mobile inverted bottleneck block (MBConv) \cite{mobilenetv2} has a long and successful history in efficient backbones \cite{efficientnet,mobilenetv2,mobilenetv3} and is still used by many new approaches \cite{efficientvit,CoatNet,efficientnetv2}. It consists of two pointwise convolutions (PWConv) and a depthwise (DWConv) in between (see \Cref{fig:mbconv} for a depiction of it). The PWConvs increase and decrease the channel dimension by the attributed expansion factor. As we concluded however in \Cref{subsec:grouping}, depthwise convolutions are not particularly hardware-efficient.
With the goal of improving hardware-efficiency, we therefore remove the depthwise convolution by utilizing the fused MBConv \cite{fusedmbconv}, which fuses the first PWConv with the DWConv into a standard convolution. Even though the fused MBConv is more hardware-efficient (as it features no depthwise convolution), it also has a significantly higher amount of MACs, although it has one layer less.
Depthwise convolutions scale better with high channel dimension, because its amount of computations is linear in regard to channel dimension, wherefore will compare the execution time of both blocks under different configurations.

In \Cref{fig:fusedexp} we compare fused and unfused MBConvs with an expansion factor of four and of six (first and second row). 
We omit the scenarios where channel numbers range from 16 to 64 and resolutions from 7 to 28, as they are not relevant for backbone architectures and GPU utilization is too low for the results to have significance.
In both figures we depict the average execution time on GPU (left side) and amount of MACs (right side) of the fused MBConv divided by the unfused one. We do this for various configurations of resolution and input channel dimension. We measure average execution time with a batch size of 200 and run for 100 iterations, the same way we measure throughput in \Cref{subsec:speed_comparison_results}.
It can be seen in \Cref{fig:fusedexp} that resolution and channel dimension have a big influence on the relative execution time and even though the fused MBConv always has more MACs (values over one), it is faster in many scenarios (values smaller than one in left part).
Only for high number of channels (over 512) the relative execution time worsens and approaches the relative MACs again.
We also evaluate the GPU latency in the same way in the supplementary material.

In \Cref{subsec:ablation} we additionally demonstrate this affect with our model LowFormer-B1.

\begin{table}[t]
    \centering
    \resizebox{8cm}{!}{
    \begin{tabular}{c|>{\columncolor[gray]{0.9}}c|c?c|c|c} 
    \Xhline{1.5pt}
        
         \multirow{2}{*}{Scenario} & Resolution  & Channels & Relative  & Relative  & Relative  \\
        
          & (pixel) & (\#)& Throughput & Latency & MACs \\ \Xhline{1.5pt}

        \multirow{2}{*}{\#1}  & 224 & 24  &  0.3 & 2.7 & 1.0  \\ \cline{2-2}
       &  28 & 196  &   \textbf{3.3}& \textbf{0.37}  & 1.0   \\   \Xhline{1.5pt}

        \multirow{2}{*}{\#2}  & 224 & 48   & 0.5 & 1.88 &1.0  \\   \cline{2-2}
       &  112 &96   &  \textbf{1.9} & \textbf{0.53 }& 1.0  \\  \Xhline{1.5pt}
        
        \multirow{2}{*}{\#3}  & 56 & 96  & 0.5 & 1.5 & 1.0  \\  \cline{2-2}
       &  14 & 384  &  \textbf{2.2}& \textbf{0.67} & 1.0  \\  \Xhline{1.5pt}
        
        \multirow{2}{*}{\#4}  & 112 & 96  & 0.6 & 1.91 & 1.0  \\   \cline{2-2}
        & 28 & 384  &   \textbf{1.8} & \textbf{0.52 }& 1.0   \\   \Xhline{1.5pt}
        
        \multirow{2}{*}{\#5}  & 224 & 96  & 0.5 & 2.16 & 1.0  \\  \cline{2-2}
       &  56 & 384  &   \textbf{1.9} & \textbf{0.46} & 1.0   \\    \Xhline{1.5pt}
        
        \multirow{2}{*}{\#6}  & 112 & 24  & 0.3 & 2.3  & 1.0  \\  \cline{2-2}
       &  14 & 196  &   \textbf{3.3} & \textbf{0.44} & 1.0  \\    \Xhline{1.5pt}

           \multirow{2}{*}{\#7}  & 224 & 48  & 0.5 & 1.15  & 1.0   \\  \cline{2-2}
         &  56 & 196  &   \textbf{2.0} & \textbf{0.87} & 1.0   \\      \Xhline{1.5pt}
         
    \end{tabular}
    }
    \caption{   Analysis of the impact of resolution on execution time relative to amount of MACs. Each scenario contains two configurations of convolutions (first row and second row in each scenario), that approximately feature the same amount of MACs (see 6th column), but strongly differ in operating resolution and number of channels. We set their throughput and latency on GPU in relation to each other(see 4th and 5th column). 
    The table shows, that convolutions operating on a higher resolution tend to be slower than lower resolution convolutions with the same amount of MACs. }
    \label{tab:highreshighchan}

\end{table}

\subsection{High Resolution vs. High Channel}
\label{subsec:highreshighchan}
Operating resolution and the number of channels of a layer have a big influence on how hardware efficient the execution is. In \Cref{tab:highreshighchan} we compare different layer configurations by creating models that consist of 20 times the same layer stacked after another. We feature seven scenarios in  \Cref{tab:highreshighchan}, each contrasting two layers with the same amount of MACs to one decimal place, but differing in channel dimension and operating resolution.
To compare the hardware-efficiency of the layers, we measure GPU latency and throughput. The models only contain standard convolutions (ungrouped).

\Cref{tab:highreshighchan} clearly shows how a higher resolution results in a less hardware efficient execution, while a higher number of channels poses a minor problem. In scenario \#1 for example the first convolution has a third of the throughput of the second one and almost thrice the latency. It operates on eight times the resolution, however features less channels and its MACs equal the second layer.
The same effect also occurs with a smaller resolution difference, as scenario \#2 shows, where the first convolution runs on twice the resolution, but still fails to execute its MACs as fast as the second one in terms of throughput and latency.

Operating on a high resolution can slow down a model more than the MACs might suggest, while reducing the operating resolution can lead to an considerably increased hardware-efficiency.
We emphasize that model scaling should be more influenced by actual measured speed \cite{efficientnet}, as MACs can be misleading, especially when scaling models by higher input resolution \cite{shvit,efficientvit}.

\begin{figure*}[hbt!]
    \includegraphics[width=\linewidth]{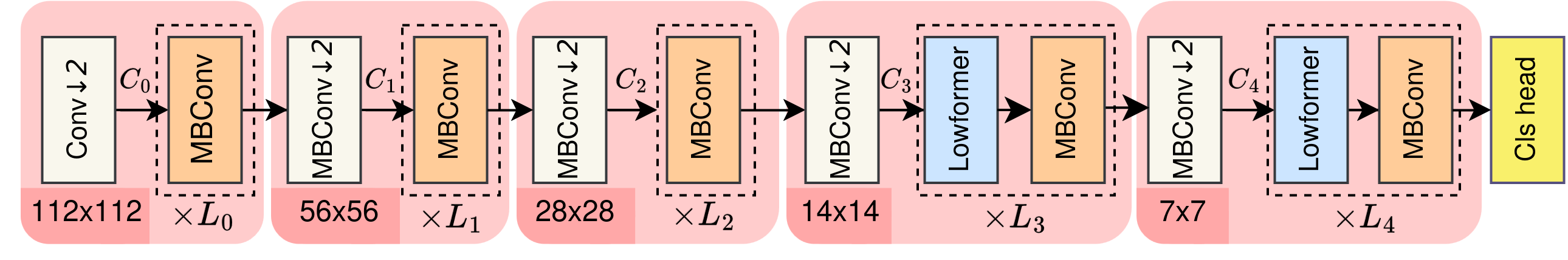}
  \caption{Architecture of LowFormer. The resolutions 
refer to a 224x224 sized input. LowFormer block can be seen in  \Cref{fig:lowformerblock}. MBConv means the mobile inverted bottleneck block, Conv means convolution and Cls head refers to the image classification head. Specification of $C_0 - C_4$ and $L_0 - L_4$ can be found in \Cref{tab:archnumbers}. }
  \label{fig:arch}
  \vspace{-0.7cm}
\end{figure*}

\section{Methodology}
\label{sec:method}

In the following we describe our proposed lightweight adaptation of the original Multi-Head Self-Attention \cite{attentionisallyouneed} and our hardware-efficient macro design.
We also explain how both arose from the insights gained from the execution time analysis in \Cref{sec:speedexperiments}.

\begin{figure*}[hbt!]
  \centering

    \begin{minipage}[b]{0.7\textwidth}

    \includegraphics[width=\textwidth]{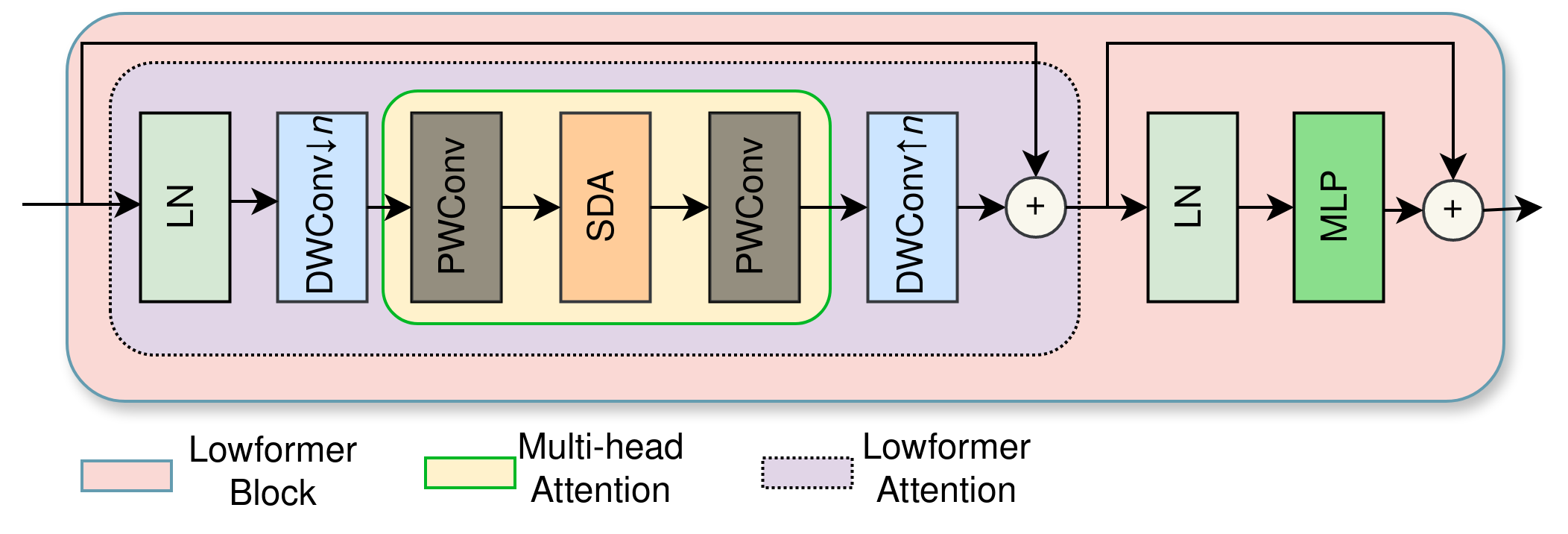}
    \caption{Lowformer block design. DWConv, PWConv, LN, MLP and SDA mean depthwise convolution, pointwise convolution, layer normalization, multi-layer perceptron and Scaled Dot-Product Attention respectively. 
  In contrast to the traditional MHSA, we encapsulate the SDA with two depthwise convolutions (the second is a transposed depthwise convolution). The projections for MHSA are realized with pointwise convolutions. The  $DW\downarrow_n$ means that the resolution is downscaled by the factor $n$ and $DW\uparrow_n$ that it is upscaled by $n$.}
  \label{fig:lowformerblock}
  \end{minipage} \hfill \begin{minipage}[b]{0.25\textwidth}
\includegraphics[width=0.85\textwidth]{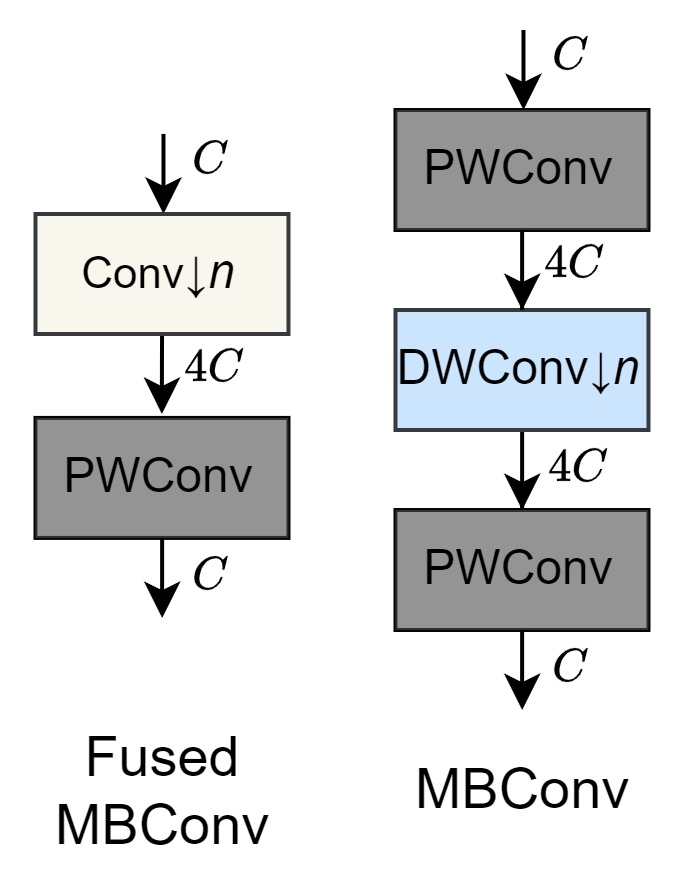}
    \caption{Structure of the fused and unfused MBConv block. $C$ refers to the channel dimension. In Conv and DWConv the "↓n" refers to a potential stride. Both have an expansion factor of 4 in this figure. }
    \label{fig:mbconv}
  \end{minipage} 
\end{figure*}

\subsection{Lightweight Attention}
\label{subsec:microdesign}
In the last two stages we employ a lightweight adaptation of the original MHSA \cite{attentionisallyouneed}, depicted in \Cref{fig:lowformerblock}. We name it LowFormer Attention. 
In our adaptation the Scaled Dot-Product Attention (SDA) is encapsulated by two depthwise convolutions and two pointwise convolutions, which 
 perform the input and output projections of the queries (Q), keys (K) and values (V). 
\paragraph{Channel compression.}
During the input projection the channel dimension of  Q, K and V is halved. After the SDA it is restored to the original channel dimension by the output projection. 
\paragraph{Lower resolution.}
In \Cref{subsec:highreshighchan} we learned how a high operating resolution slows down convolutions disproportionately. We transferred that insight to attention, wherefore convolutions (see \Cref{fig:lowformerblock}) also down- and upsample the resolution of the feature maps around the SDA in the forelast stage, such that the attention operation is executed on half the resolution. In \Cref{subsec:ablation} we show that this measure has a substantial effect on throughput and latency.

Due to these two adaptations all dimensions of the input to the SDA (channel dimension, height and width) are reduced, wherefore we connote LowFormer Attention with the attribute lightweight.

\paragraph{MLP following Attention.}
Following \citeauthor{attentionisallyouneed}\cite{attentionisallyouneed}, we append layer normalization and a multi-layer perceptron (MLP) after the LowFormer Attention. We found that its effect on model accuracy is significant and \citeauthor{efficientvitbad}\cite{efficientvitbad} pointed out, the MLP is more hardware-efficient than the attention operation. 

\subsection{Macro Design}
\label{subsec:macrodesign}
Our macro design is based on EfficientViT \cite{efficientvit} and MobileViT \cite{mobilevit}. We feature five stages and adapted the architecture mainly according to the following two recipes, that are fueled by the insights gained from the execution time analysis in \Cref{sec:speedexperiments}. 
The whole architecture is depicted in \Cref{fig:arch}. We present five different versions of our architecture, namely B0-B3 (B0, B1, B1.5, B2, B3). Architecture details are listed in  \Cref{tab:archnumbers}.

\paragraph{Less layers in the first stages.}
From the insights in \Cref{subsec:highreshighchan} we conclude that a minimal amount of layers in the first stages is more hardware-efficient (see \Cref{tab:archnumbers}).
It proved optimal to apply the reduction for the first three stages. Most computation is therefore concentrated in the last two stages, where for an input size of 224$\times$224, the operating resolutions are 14$\times$14 and 7$\times$7.


\paragraph{Fusing Depthwise and Pointwise Convolutions.}
In \Cref{subsec:grouping} we showed that depthwise convolutions are not as hardware-efficient as standard convolution and in \Cref{subsec:fusedunfused} we came to the conclusion that the fused MBConv (see \Cref{fig:mbconv}) can be faster than the unfused one. This effect diminishes however with increasing number of channels. We therefore fused the MBConv in our architecture, wherever the number of input channels reach at most 256, except for the stride 2 MBConv block in the last stage, which we fuse nevertheless (see \Cref{fig:arch}).
We additionally fuse the depthwise and pointwise convolutions after the SDA in the LowFormer Attention (see \Cref{fig:lowformerblock}).
We confirm the effect of this approach by reverting the fusion of the MBConv block for LowFormer-B1 in the ablation in \Cref{subsec:ablation}.

\begin{table}[hbt!]
    \centering
    \resizebox{8cm}{!}{
    \begin{tabular}{c|c|c}
        \Xhline{1.5pt}
         Model  & $\{L_0,L_1,L_2,L_3,L_4 \}$  & $\{C_0,C_1,C_2,C_3,C_4 \}$  \\
         \Xhline{1.5pt}

         LowFormer-B0 & $\{0,0,0,3,4\}$ & $\{16,32,64,128,256\}$  \\
         LowFormer-B1 & $\{0,0,0,5,5\}$ & $\{16,32,64,128,256\}$  \\
        LowFormer-B1.5 & $\{0,0,0,6,6\}$ & $\{20,40,80,160,320\}$  \\
         
         LowFormer-B2 & $\{0,0,0,6,6\}$ & $\{24,48,96,192,384\}$  \\
         LowFormer-B3 & $\{1,1,2,6,6\}$ & $\{32,64,128,256,512\}$  \\
        
        \Xhline{1.5pt}
     
    \end{tabular}
    }
    \caption{Specification of LowFormer architecture versions B0-B3. The number of layers ($L_0-L_4$) and channels ($C_0-C_4$) relates to \Cref{fig:arch}. }
    \label{tab:archnumbers}
\vspace{-0.4cm}
\end{table}

\begin{table}[hbt!]
    \centering
    \resizebox{8cm}{!}{
    \begin{tabular}{c|c|c|c}
        \Xhline{1.5pt}
         \multirow{2}{*}{Model}  &  Mobile GPU Latency & ARM CPU Latency &  Top-1   \\
         & (ms) & (s) & (\%) \\
         \Xhline{1.0pt}

        EfficientFormerV2-S0 \cite{efficientformerv2}  &   26 & 1.201 & 73.7 \\
       
        MobileOne-S3 \cite{efficientvit}  &   34 & 1.20 &78.1 \\

        LowFormer-B0 (ours)  &  \textbf{20} & \textbf{0.835} & \textbf{78.4} \\
        
        \Xhline{1.0pt}

        EfficientFormerV2-S1 \cite{efficientformerv2}  &   35 & 1.388 & 77.9 \\
        ResNet-50 \cite{resnetpaper}  &   58 &  2.376 &  79.0  \\
        
        RepViT-M1.1 \cite{repvit}  &   39 & 1.346 &79.4 \\
        
        MobileOne-S4 \cite{efficientvit}  &   45 & 1.889 &  79.4 \\
        
        LowFormer-B1 (ours)  &  \textbf{34} & \textbf{1.186} & \textbf{79.9} \\
        
         \Xhline{1.0pt}

        EfficientFormerV2-S2 \cite{efficientformerv2}  &   \textbf{54} & 2.061 & 80.4 \\
        RepViT-M1.5 \cite{repvit}  &   \textbf{54} & 2.315 & \textbf{81.2} \\ 

        LowFormer-B1.5 (ours)  &  58 & \textbf{1.970} & \textbf{81.2 }\\
        
        
          \Xhline{1.5pt}





     
    \end{tabular}
    }
    \caption{Mobile GPU and ARM CPU Latency of backbones. Evaluation resolution is $224\times224$ and batch size is set to 1.}
    \label{tab:mobilelatency}
    \vspace{-0.3cm}
\end{table}


\section{Experiments}

\begin{table*}[hbt!]
    \centering
   \resizebox{14cm}{!}{
    \begin{tabular}{cccccccc}
        \Xhline{1.5pt}
         \multirow{2}{*}{Model} & \multirow{2}{*}{Venue} & Params  & MACs & GPU Throughput & GPU Latency & Resolution & Top-1  \\
         & & (M) & (M) & (images/s) & (ms) & (pixel) & (\%)  \\
        \Xhline{1.5pt}
         
         MobileViG-Ti \cite{mobilevig} & CVPRW 2023 & 5.3 & 661 & 2500 & 0.39 & 224 &  75.7  \\
        PVTv2-B0 \cite{pvtv2} & CVM 2022 & 3.4 & 566 & 2164 & 0.50 & 224 &  70.5 \\

        EdgeViT-XXS \cite{edgevit} & ECCV 2022 & 4.1 & 551 & 2816  & 0.41 & 224 &  74.4   \\
        
         GhostNetV2 x1.0 \cite{ghostnetv2} & NeurIPS 2022 & 6.2 & 183 & 375 & 1.93 & 224 &  75.3 \\
        
         FastViT-T8 \cite{fastvit} & CVPR 2023 & 3.6 & 690 & 1694 & 0.54 & 256 &  75.6  \\

         EfficientMod-xxs \cite{efficientmodulation} & ICLR 2024 & 4.7 & 579 & 2857 & 0.39 & 224 &  76.0    \\
        
         EfficientViT-M5 \cite{efficientvitbad}& CVPR 2023  & 12.4 & 521 & 5681 & 0.30 & 224 &  77.1    \\
        RepViT-M0.9 \cite{repvit} & CVPR 2024 & 5.1 & 816 & 2512 & 0.47 & 224 &  77.4   \\
       SHViT-S3 \cite{shvit} & CVPR 2024 & 14.3 & 601 & \underline{5780}  & 0.28 & 224 &  77.4   \\
        MobileOne-S2 \cite{mobileone} & CVPR 2023 & 7.8 & 1298 & 2967 & 0.35 & 224 &  77.4  \\

         \rowcolor{lightergray} LowFormer-B0 (ours)& &  14.1 & 944 & \textbf{5988} & 0.30 & 224 &  \textbf{78.4 }  \\

         \Xhline{1.5pt}

         EdgeViT-XS \cite{edgevit} & ECCV 2022 & 6.8 & 1127 & 2127  & 0.56 & 224 &  77.5  \\
         MobileOne-S3 \cite{mobileone} & CVPR 2023 & 10.1 & 1895 & 2433 & 0.48 & 224 &  78.1 \\
         MobileViG-S \cite{mobilevig} &  CVPRW 2023 & 7.3 & 983 & 1724 & 0.57 & 224 &  78.2  \\
        
        EfficientMod-xs \cite{efficientmodulation} & ICLR 2024 & 6.6 & 773 & 2352 & 0.50 & 224 &  78.3   \\
            PVTv2-B1 \cite{pvtv2} & CVM 2022 & 13.1 & 2108 & 1228 & 0.98 & 224 &  78.7  \\

EfficientViT-B1 \cite{efficientvit} & ICCV 2023 & 9.1 & 519 & 2739 & 0.44 & 224 &  79.4   \\
      
              SHViT-S4 \cite{shvit} & CVPR 2024 & 16.5 & 986 & \textbf{4255}  & 0.35 & 256 &  79.4   \\

         \rowcolor{lightergray} LowFormer-B1 (ours) & & 17.9 & 1410 & \underline{4237} & 0.43 & 224 &  \textbf{79.9 }  \\

         \Xhline{1.5pt}

        FAT-B0 \cite{fat} & NeurIPS 2023& 4.5 & 754 & 1620  & 0.70 & 224 &  77.6   \\
        MobileOne-S4 \cite{mobileone} & CVPR 2023 & 14.8 & 297 & 1550 & 0.82 & 224 &  79.4  \\
        
         RepViT-M1.1 \cite{repvit} & CVPR 2024 & 8.2 & 1338 & 1941 & 0.61 & 224 &  79.4    \\
         EfficientViT-M5 \cite{efficientvitbad}& CVPR 2023 & 12.4 & 1494 & \underline{2247} & 0.51 & 384 &  79.8    \\
         \rowcolor{lightergray} LowFormer-B1 (ours) & & 17.9 & 1843 & \textbf{3378} & 0.48 & 256 &  \textbf{80.2}   \\
 
 \Xhline{1.5pt}

        
      FAT-B1 \cite{fat} & NeurIPS 2023 & 7.8 & 1243 & 1204  & 0.96 & 224 &  80.1  \\
        
        FastViT-SA12 \cite{fastvit} & CVPR 2023& 10.9 & 1943 & 1075 & 0.95 & 256 &  80.6   \\
       SHViT-S4 \cite{shvit} & CVPR 2024 & 16.5 & 2224 & \underline{1785}  & 0.62 & 384 &  81.0 \\
      EdgeViT-S \cite{edgevit} & ECCV 2022 & 11.1 & 1910 & 1449  & 0.85 & 224 &  81.0  \\

        \rowcolor{lightergray} LowFormer-B1.5 (ours) & & 33.9 & 2573 & \textbf{2739} & 0.66 & 224 &  \textbf{81.2}  \\
         
        \Xhline{1.5pt}

        EfficientViT-M5 \cite{efficientvitbad}& CVPR 2023 & 12.4 & 2799 & 1197 & 0.92 & 512 &  80.8    \\
        EfficientMod-s \cite{efficientmodulation} & ICLR 2024 & 12.9 & 1402 & \underline{1381} & 0.83 & 224 &  81.0   \\
    
        RepViT-M1.5 \cite{repvit} & CVPR 2024 & 14.0 & 2276 & 1146 & 1.03 & 224 &  81.2  \\
        FFNet-1 \cite{ffnet} & arXiv 2024 & 13.8 & 3000 & 1090 & 0.96 & 256 &  81.3 \\
         \rowcolor{lightergray} LowFormer-B2 (ours) & & 45.0 & 3689 & \textbf{2227} & 0.88 & 224 &  \textbf{81.6}  \\

        \Xhline{1.5pt} 
         
         SHViT-S4 \cite{shvit} & CVPR 2024 & 16.5 & 3971 & 1160  & 1.04 & 512 &  82.0 \\
        EfficientViT-B2 \cite{efficientvit} & ICCV 2023 & 15.0 & 1584 & \underline{1298} & 0.92 & 224 &  82.1  \\ 
        RepViT-M2.3 \cite{repvit}& CVPR 2024 & 22.9 & 4520 & 642 & 1.84 & 224 &  82.5    \\
       FastViT-SA24 \cite{fastvit}& CVPR 2023 & 10.9 & 3769 & 606 & 1.70 & 256 & 82.6   \\

        MobileViG-B \cite{mobilevig} & CVPRW 2023 & 26.7 & 2792 & 869 & 1.24 & 224 &  82.6  \\
       
        \rowcolor{lightergray}  LowFormer-B3 (ours)& & 57.1 & 4479 & \textbf{1562} & 1.24 & 192 &  \textbf{82.7} \\
         
         \Xhline{1.5pt}

        FFNet-2 \cite{ffnet} & arXiv 2024 & 26.9 & 6060 & 496 & 2.12 & 256 &  82.9  \\
       iFormer-S \cite{inceptiontransformer}& NeurIPS 2022 & 19.9 & 4825 & 555  & 2.11 & 224 &  83.4  \\

          EfficientViT-B3 \cite{efficientvit} & ICCV 2023& 39.0 & 3953 & \underline{666} & 1.88 & 224 &  83.4   \\ 
       
       FAT-B3 \cite{fat}& NeurIPS 2023 & 29.0 & 4631 & 402  & 2.82 & 224 &  83.6   \\
     
       FastViT-SA36 \cite{fastvit} & CVPR 2023 & 30.4 & 5595 & 429 &  2.00 & 256 & 83.6    \\
       
       \rowcolor{lightergray} LowFormer-B3 (ours)& & 57.1 & 6098 & \textbf{1162} & 1.55 & 224 &  \textbf{83.6}  \\

        \Xhline{1.5pt}



         

    \end{tabular}
    }
    \caption{Performance on ImageNet-1K validation set. Neither distillation nor pretraining is used for fair comparison. The table is divided in different complexity classes, determined by throughput. Results in \textbf{bold} are the best results for each complexity class, while underlined results refer to the second best.}
    \label{tab:imagenetresults}
    \vspace{-0.4cm}
\end{table*}

\label{sec:Experiments}

\subsection{ImageNet-1K Classification}
\label{subsec:imagenetclass}

\paragraph{Settings.}
We conduct image classification experiments on ImageNet-1K \cite{imagenet}, which includes 1.28M training and 50K validation images for 1000 categories. All models were trained from scratch using mostly the same setting as \citeauthor{efficientvit}\cite{efficientvit} and featuring an input resolution of 224. We also trained for a total of 320 epochs using AdamW \cite{adamwpaper} optimizer and a learning rate of $10^{-3}$, however we use a batch size of 512. 
As learning rate scheduler we use cosine decay \cite{cosinelrpaper} and 20 warm-up epochs with a linear schedule. We also feature the multi-scale learning from \citeauthor{efficientvit}\cite{efficientvit}.
We trained LowFormer-B3 with a batch size of 2400 and a base learning rate of $3\times10^{-3}$. For LowFormer-B2 we had a batch size of 850 and a base learning rate of $8.3\times10^{-4}$.

\paragraph{GPU Throughput and Latency.}
\label{subsec:speed_comparison_results}
In  \Cref{tab:imagenetresults} we measure speed by GPU throughput and latency.
For GPU throughput we run for 100 iterations on a Nvidia A40 graphic card with a batch size of 200 and take the median time per input image to calculate the total throughput.
To measure GPU latency we run for 400 iterations on a Nvidia Titan RTX, with a batch size of 16. For the latter we also compile all models to TorchScript code and optimize them for inference \footnote{\url{https://pytorch.org/docs/stable/generated/torch.jit.optimize_for_inference.html}}.
We feature 5 warm-up iterations for latency and throughput measurement.

\paragraph{Results.}
For a better comparison between competing approaches we include a version of LowFormer-B1 and LowFormer-B3 that are evaluated on a different resolution (see \Cref{tab:imagenetresults}).
The variants of LowFormer outperform all other approaches (see \Cref{tab:imagenetresults}) with a similar or lower GPU throughput in regard to top-1 accuracy. 
LowFormer-B1$_{r256}$, tested on resolution 256, has a 0.8\% higher top-1 accuracy than EfficientViT-B1 \cite{efficientvit} and a 23\% higher throughput. It also similarly outperforms RepViT-M1.1 \cite{repvit}, but has a 73\% higher throughput and a 21\% lower latency. LowFormer-B1, tested on resolution 224, only achieves a 0.44 \% increase in accuracy compared to SHViT-S4$_{r256}$ \cite{shvit}, with a similar throughput. The SHViT architecture however only increases their model complexity in an inefficient way (see \Cref{subsec:highreshighchan}) by increasing input resolution, wherefore their best model SHViT-S4$_{r512}$ (tested on resolution 512) scores 1.64 \% worse in top-1 accuracy than LowFormer-B3 with a similar throughput.
The hardware-efficiency of our design and the inconsistent relationship between amount of MACs and execution time becomes especially apparent, when comparing LowFormer-B3 to GhostNetV2x1.0 \cite{ghostnetv2}. Former has 3$\times$ the throughput, 20\% lower latency, 36 $\times$ the amount of MACs and scores 8.3\% better in top-1 accuracy.

\begin{figure}
  \includegraphics[width=\linewidth]{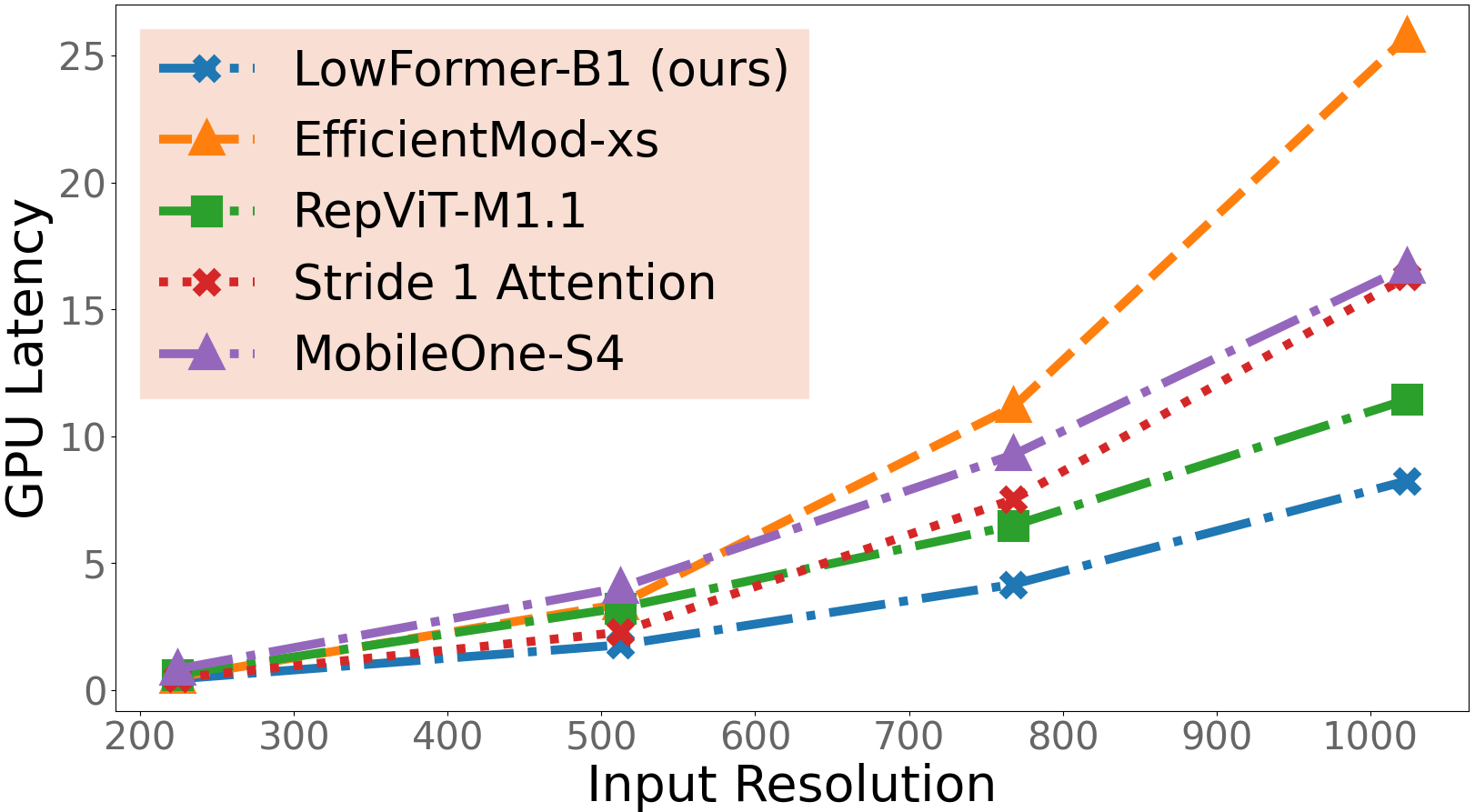}
  \caption{Impact of input resolution on GPU Latency for LowFormer-B1 (ours), LowFormer-B1 without downsampling in LowFormer Attention (see \Cref{subsec:ablation}), EfficientMod-xs \cite{efficientmodulation}, FastViT-T8 \cite{fastvit} and RepViT-M1.1 \cite{repvit}.}
  \label{fig:resolution_scaling}
  \vspace{-0.2cm}
\end{figure}

\paragraph{Resolution Scaling.}
In \Cref{fig:resolution_scaling} we investigate for LowFormer-B1 and other approaches the impact of increasing the input resolution on GPU latency. 
LowFormer-B1 outperforms depicted models in top-1 accuracy on ImageNet-1K and at the same time remains considerably faster, independent of the input resolution.
In the supplementary we further perform the same evaluation with a batch size of 200 to estimate change in throughput. 

\paragraph{Mobile GPU and CPU Latency.}
In \Cref{tab:mobilelatency} we also verify the efficiency of our model on mobile GPU (ARM Mali-G76 MP12) and ARM CPU (ARM Cortex A53). We run the models with a batch size of 1 and a resolution of 224$\times$224. For the ARM CPU we ran for 30 iterations, while for the mobile GPU we ran for 10000 iterations, with 1000 warm-up iterations.
Even though our architecture is not specifically optimized for edge application, 
LowFormer-B1 for examples scores 0.5\% better in top-1 accuracy than MobileOne-S4 \cite{mobileone}, who has a 32\% higher mobile GPU latency and a 59\% higher ARM CPU latency.

\subsection{Object Detection \& Semantic Segmentation}
We utilize the pretrained backbones and apply them in object detection and semantic segmentation.
We train and evaluate on COCO 2017 \cite{cocopaper} and ADE20K \cite{ade20k} respectively using mmdetection \cite{mmdetection} and mmsegmentation \cite{mmseg2020}.
For object detection we take the RetinaNet framework \cite{retinanet}, while for Semantic Segmentation we plug our backbone into the Semantic FPN \cite{semanticfpn}.
For backbone GPU throughput and latency measurement we follow the protocol in \Cref{subsec:imagenetclass}, but evaluate on input resolution of 512$\times$512.

\paragraph{Object Detection.}
We train the models for 12 epochs (1x schedule), following \cite{efficientvit,fat}.
Regarding results, LowFormer-B2 for example outperforms FAT-B0 \cite{fat} by \textbf{+1.0} AP, while having  93\% higher backbone throughput on resolution 512$\times$512 (see \Cref{tab:objectdetection}).   

\paragraph{Semantic Segmentation.}
For semantic segmentation we train the models for 40K iterations with a batch size of 32, following \cite{fat,repvit,fastvit,efficientmodulation}. We use AdamW optimizer \cite{adamwpaper}, cosine annealing for the learning rate \cite{cosinelrpaper} with a base learning rate of $2\times 10^{-3}$ and 1K warm-up steps with linear increase. 
LowFormer-B1 for example has 2.4$\times$ the throughput and a 20\% lower latency than EfficientFormerV2-S2\cite{efficientformerv2}, but achieves \textbf{+0.4} mIoU when plugged into Semantic FPN (see \Cref{tab:segmentation}).

\begin{table}[t]
    \centering
    \resizebox{8cm}{!}{
    \begin{tabular}{c|cc|cccccc}
        \Xhline{1.5pt}
         \multirow{2}{*}{Backbone}  & Throughput & Latency &AP &  AP$_{50}$ & AP$_{75}$ & AP$_{s}$ &AP$_{m}$ & AP$_{l}$ \\
         & (images/s) & (ms) & (\%) & (\%) & (\%) & (\%) & (\%) &  (\%)  \\
         \Xhline{1.0pt}

        MobileNetV3 \cite{mobilenetv3}  & 862 & 0.97 & 29.9 & 49.3 & 30.8 & 14.9 & 33.3  & 41.1 \\
        
        EfficientViT-M4 \cite{efficientvitbad}  & \textbf{1700}  & 0.58 & 32.7 & 52.2 & 34.1 & 17.6 & 35.3 & 46.0  \\ 
        PVTv2-B0  \cite{pvtv2} & 355 & 3.01 & 37.2 & 57.2 & 39.5 & \textbf{23.1} & 40.4 & 49.7 \\

        LowFormer-B0 (ours)  & 1190  & 1.24 & \textbf{38.6 }& \textbf{59.1} & \textbf{40.9} & 21.8 & \textbf{41.8 }& \textbf{51.7}  \\
        \Xhline{1.0pt}
        
        EdgeViT-XXS \cite{edgevit}  & 518 & 2.18 & 38.7 & 59.0 & 41.0 & \textbf{22.4} & 42.0  & 51.6   \\
        
        LowFormer-B1 (ours)  & \textbf{840} & 1.77 & \textbf{39.4 }& \textbf{59.8} & \textbf{41.7 }& \textbf{22.4} & \textbf{42.9} & \textbf{52.4}  \\
        \Xhline{1.0pt}
     
        FAT-B0 \cite{fat}  & 232 &  4.58 & 40.4 & 61.6 & 42.7 & 24.0 & 44.3 & 53.1  \\
        
        EdgeViT-XS \cite{edgevit}  & 400 & 2.89 & 40.6 & 61.3 & 43.3 & 25.2 & 43.9  & 54.6   \\
        

        PVTv2-B1  \cite{pvtv2} & 215 & 5.22 & 41.2 & 61.9 & 43.9 & \textbf{25.4} & 44.5 & 54.3  \\
       
        LowFormer-B2 (ours) & \textbf{450} & 4.42 & \textbf{41.4} & \textbf{62.2} & \textbf{ 44.1 } &  24.5 &  \textbf{45.1 }& \textbf{55.5}  \\

        \Xhline{1.0pt}

        FAT-B1 \cite{fat}  & 174 &  6.32 & 42.5 & 64.0 & 45.1 & 26.9 & 46.0 & \textbf{56.7}  \\
        
        LowFormer-B3 (ours) & \textbf{245} & 7.40 & \textbf{43.1} & \textbf{ 64.5} &  \textbf{45.9} &  \textbf{27.1} &  \textbf{47.1} & \textbf{ 56.7}  \\
     \Xhline{1.5pt}
    \end{tabular}
    }
    \caption{Comparison results on object detection on COCO 2017 \cite{cocopaper} using RetinaNet \cite{retinanet} head. Backbone throughput and latency are measured under resolution of 512$\times$512 and on GPU.}
    \label{tab:objectdetection}

\end{table}


\begin{table}[t]
    \centering
    \resizebox{8cm}{!}{
    \begin{tabular}{c|cc|c}
        \Xhline{1.5pt}
         \multirow{2}{*}{Backbone}  & GPU Throughput & GPU Latency  & mIoU    \\
         & (images/s) &  (ms) & (\%) \\
         \Xhline{1.0pt}

        ResNet50  \cite{resnetpaper} & 271 & 3.32 & 36.7 \\
        PVTv2-B0  \cite{pvtv2} & 355 & 3.01 & 37.2 \\
        FastViT-SA12  \cite{fastvit} & 265 & 3.77 & 38.0 \\
        EdgeViT-XXS \cite{edgevit}  & 518 & 2.18 & 39.7  \\
        LowFormer-B1 (ours) & \textbf{840} &  1.77 & \textbf{ 39.7} \\
        \Xhline{1.0pt}

        RepViT-M1.1  \cite{repvit} & 404 & 2.87 & 40.6 \\
        FastViT-SA24  \cite{fastvit} & 151 & 6.88 & 41.0 \\
        EdgeViT-XS \cite{edgevit}  & 400  & 2.89 & 41.4  \\
        FAT-B0 \cite{fat}  & 232 & 4.58 & 41.5 \\
                
        EfficientFormerV2-S2 \cite{efficientformerv2}  & 182 & 5.58 & 42.4 \\
        
        PVTv2-B1  \cite{pvtv2} & 215 & 5.22 & 42.5 \\

        LowFormer-B2 (ours) & \textbf{450} & 4.42 & \textbf{ 42.8} \\
        \Xhline{1.0pt}
        
        FAT-B1 \cite{fat}  & 174 & 6.32 & 42.9 \\
        RepViT-M1.5  \cite{repvit} & 238 & 5.08 & 43.6 \\
        
        FastViT-MA36  \cite{fastvit} & 86 & 13.18 & 44.6 \\
        LowFormer-B3 (ours) & \textbf{245}  & 7.40 & \textbf{ 44.6} \\
    
        \Xhline{1.5pt}
     
    \end{tabular}
    }
    \caption{Results on semantic segmentation, using Semantic FPN \cite{semanticfpn}. Backbone throughput and latency are measured under resolution of 512$\times$512.}
    \label{tab:segmentation}
    \vspace{-0.3cm}
\end{table}

\subsection{Ablation Study}
\label{subsec:ablation}
In \Cref{tab:ablation} we ablate our model design decisions. 
We revert a singular design decision of LowFormer-B1 to demonstrate the impact of that change on accuracy, throughput and latency. 
The featured ablations are the following:
\vspace{-0.2cm}
\begin{itemize}
    \item We replace all fused MBConv blocks with the unfused version.
\vspace{-0.2cm}
    \item We remove the LowFormer block  and add an additional MBConv to each stage, such that the ablated model has the same throughput as the baseline (called "attention removed" in \Cref{tab:ablation}).
    \vspace{-0.2cm}
    \item We replace our LowFormer Attention with ReLU linear attention from \citeauthor{efficientvit}\cite{efficientvit} in order to compare our attention approach with other recent adaptations.
    \vspace{-0.2cm}
    \item We omit the downscaling of the feature maps for the LowFormer Attention.

\end{itemize}

\paragraph{Discussion.}
Replacing the fused MBConv with the unfused one results in a 0.8\% lower top-1 accuracy, while its GPU throughput is also 16\% lower (see \Cref{tab:ablation}). 
As we can see, next to an improved throughput, fusing the MBConv can increase performance significantly. \\
On the other side, replacing the LowFormer block with convolutional layers in every stage only results in a smaller drop in top-1 accuracy. Because of the additional layers however, the potential to upscale the depth of the architecture is reduced, as the more layers a stage has, the less effective they deliberately become (see \citeauthor{efficientnet}\cite{efficientnet}).\\
When we remove the downsampling for the LowFormer Attention, top-1 accuracy stays the same, but GPU throughput and latency worsen. In \Cref{fig:resolution_scaling} we can see that for higher input resolutions the latency difference multiplies. For input resolution 1024$\times$1024 for example, latency is increased by 70\%.

\begin{table}[t]
    \centering
    \resizebox{8cm}{!}{
    \begin{tabular}{c|cc|cc|c}
        \Xhline{1.5pt}
         \multirow{2}{*}{Model version}  & Params &  MACs & GPU Throughput & GPU Latency & Top-1   \\
         & (M) & (M) & (images/s) & (ms) & (\%)  \\
         \Xhline{1.0pt}
            
        unfused MBConv & 12.4 & 716 & 3558 {\scriptsize  \textbf{(-16\%)}} & 0.43 & 79.1 {\scriptsize \textbf{(-0.8)}}  \\
        attention removed & 14.46 & 1643 & 4098 {\scriptsize  \textbf{(-3\%)}}  & 0.41 & 79.7 {\scriptsize \textbf{(-0.2)}} \\
        relulinear att & 14.15 & 1210 & 3367 {\scriptsize  \textbf{(-20\%)}} & 0.49 & 79.6 {\scriptsize \textbf{(-0.3)}} \\
        high-res attention & 17.65 & 1494 & 3759 {\scriptsize  \textbf{(-11\%)}} & 0.47 & \textbf{79.9} {\scriptsize \textbf{(-0.0)}}\\
        \Xhline{1.0pt}
        Baseline (B1) & 17.94 & 1410 &  \textbf{4237} & 0.43 & \textbf{79.9}  \\
        
        \Xhline{1.5pt}
     
    \end{tabular}
    }
    \caption{Ablation study of LowFormer-B1, featuring singular changes to the original model and putting them in relation to the original model. Bold entries mark the best in its column.}
    \label{tab:ablation}

\end{table}

\section{Conclusion}

We have shown how a high resolution and depthwise convolutions negatively impact hardware-efficiency and proposed a recipe on how depthwise convolutions can be replaced in an architecture. We also proposed a simple lightweight attention and demonstrated that letting it operate on a lower resolution does not reduce accuracy, but leads to a considerably faster execution, especially when the input resolution of the model is increased. Our hardware-efficient macro and micro design yields significant speed-ups compared to previous approaches.
We further proved the applicability of our backbone architecture to object detection and semantic segmentation.



\section*{Acknowledgements}
This work was funded by European Union-NextGenerationEU under Project PRIN 2022 EXTRA EYE and Project PRIN 2022 PNRR TEAM.


{\small
\bibliographystyle{ieeenat_fullname}
\bibliography{PaperForReview}
}

\end{document}


\title{Supplementary Material \\ LowFormer: Hardware Efficient Design for Convolutional Transformer
Backbones}

\author{Moritz Nottebaum\textsuperscript{1}\\
{\tt\small nottebaum.moritz@spes.uniud.it}
\and Matteo Dunnhofer\textsuperscript{1}\\
{\tt\small matteo.dunnhofer@uniud.it}
\and Christian Micheloni\textsuperscript{1}\\
{\tt\small christian.micheloni@uniud.it}
\\
\textsuperscript{1}University of Udine, Italy\\
}

\maketitle

\begin{figure}
\centering

\includegraphics[width=\linewidth]{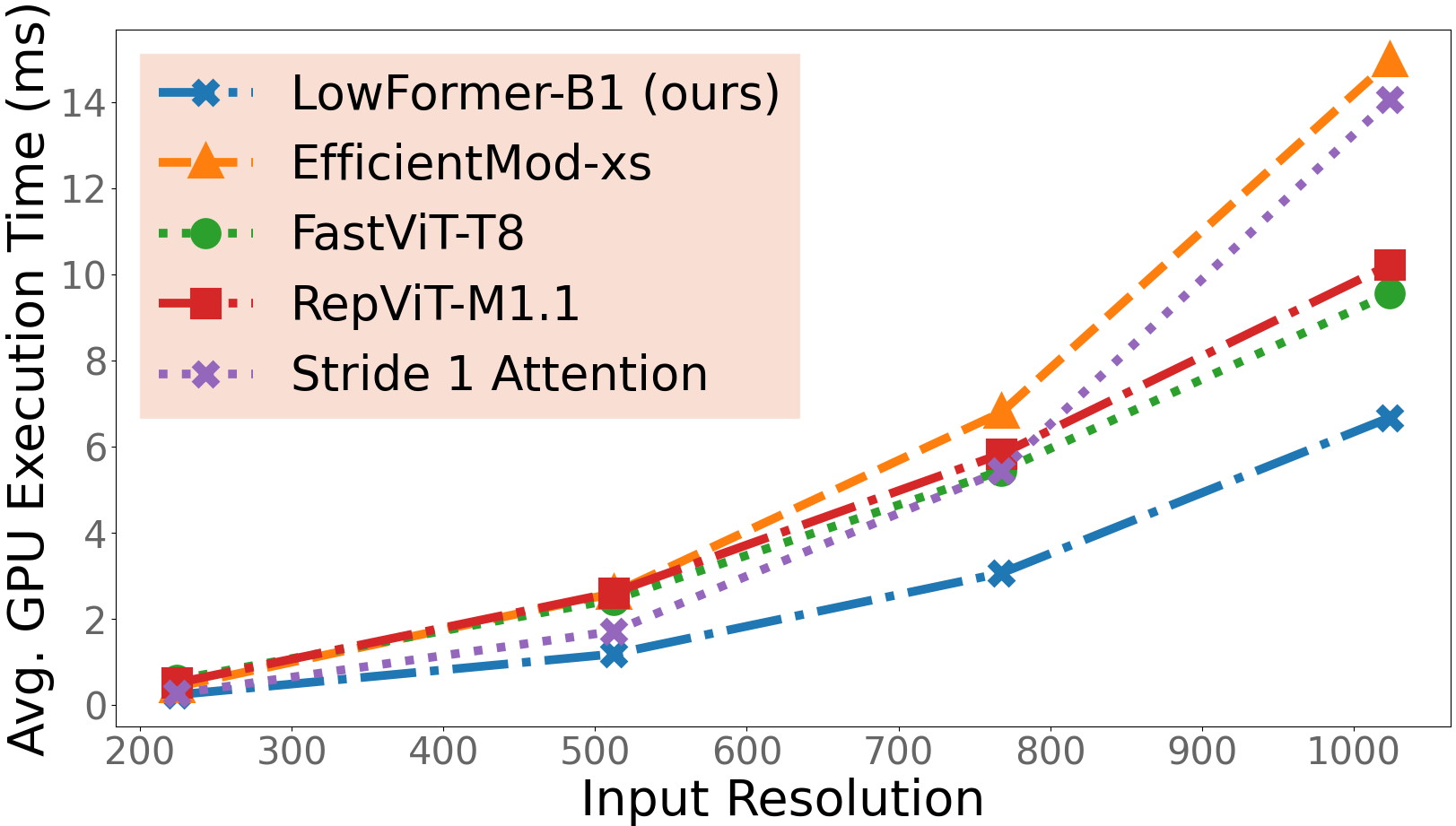}
  \caption{ Average execution time for different models, depending on resolutions. The execution time is measured with batch size 200 and 5 warm-up iterations followed by 100 iterations from which the median is taken.}
  \label{fig:avgexeresolution}
\end{figure}

\section{Latency Evaluation of fused vs unfused MBConv}

\begin{figure}
\centering
MBConv with expansion factor 4

\vspace{0.1cm}

\includegraphics[width=\linewidth]{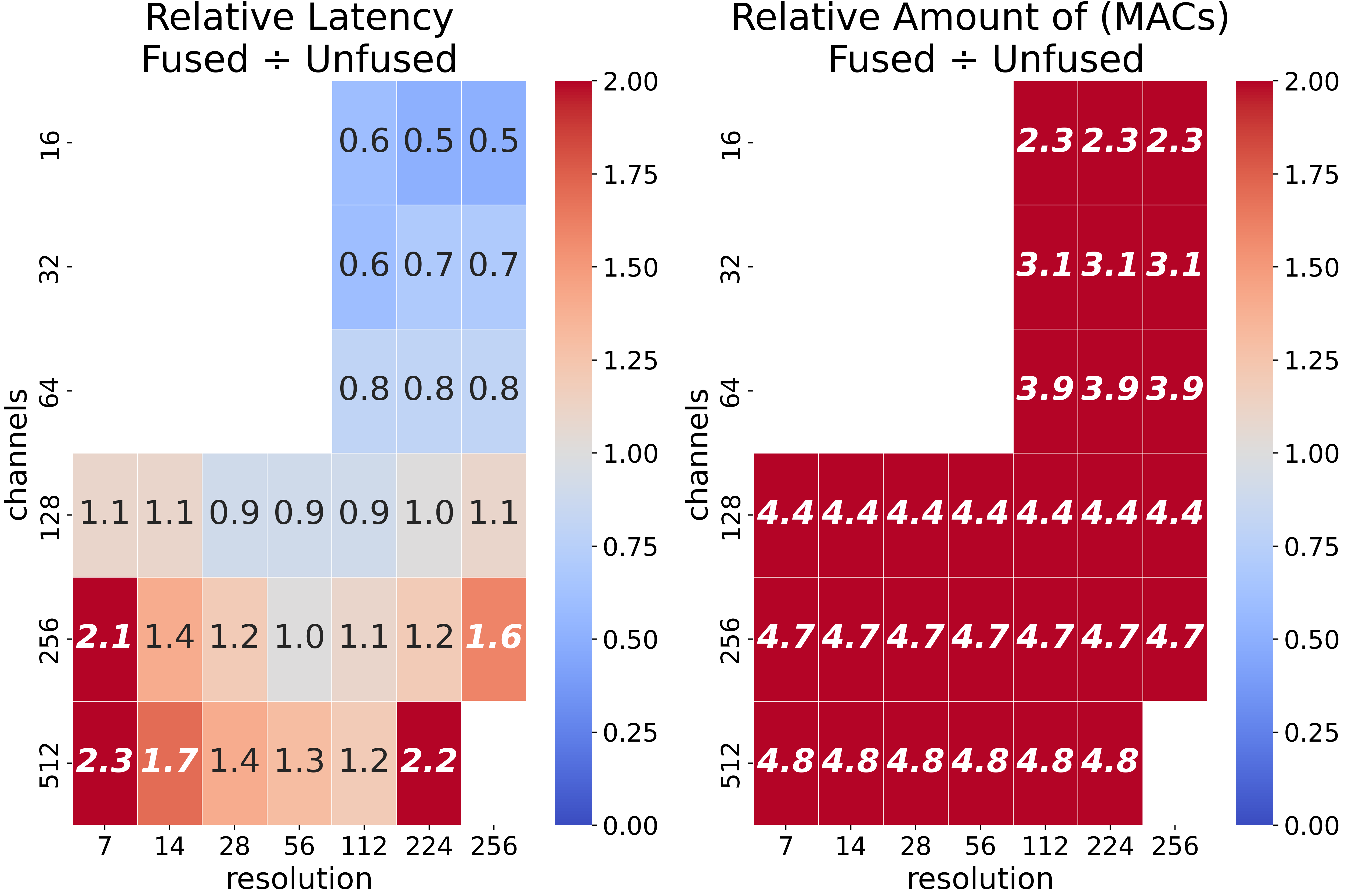}

\vspace{0.3cm}

MBConv with expansion factor 6

\vspace{0.1cm}

\includegraphics[width=\linewidth]{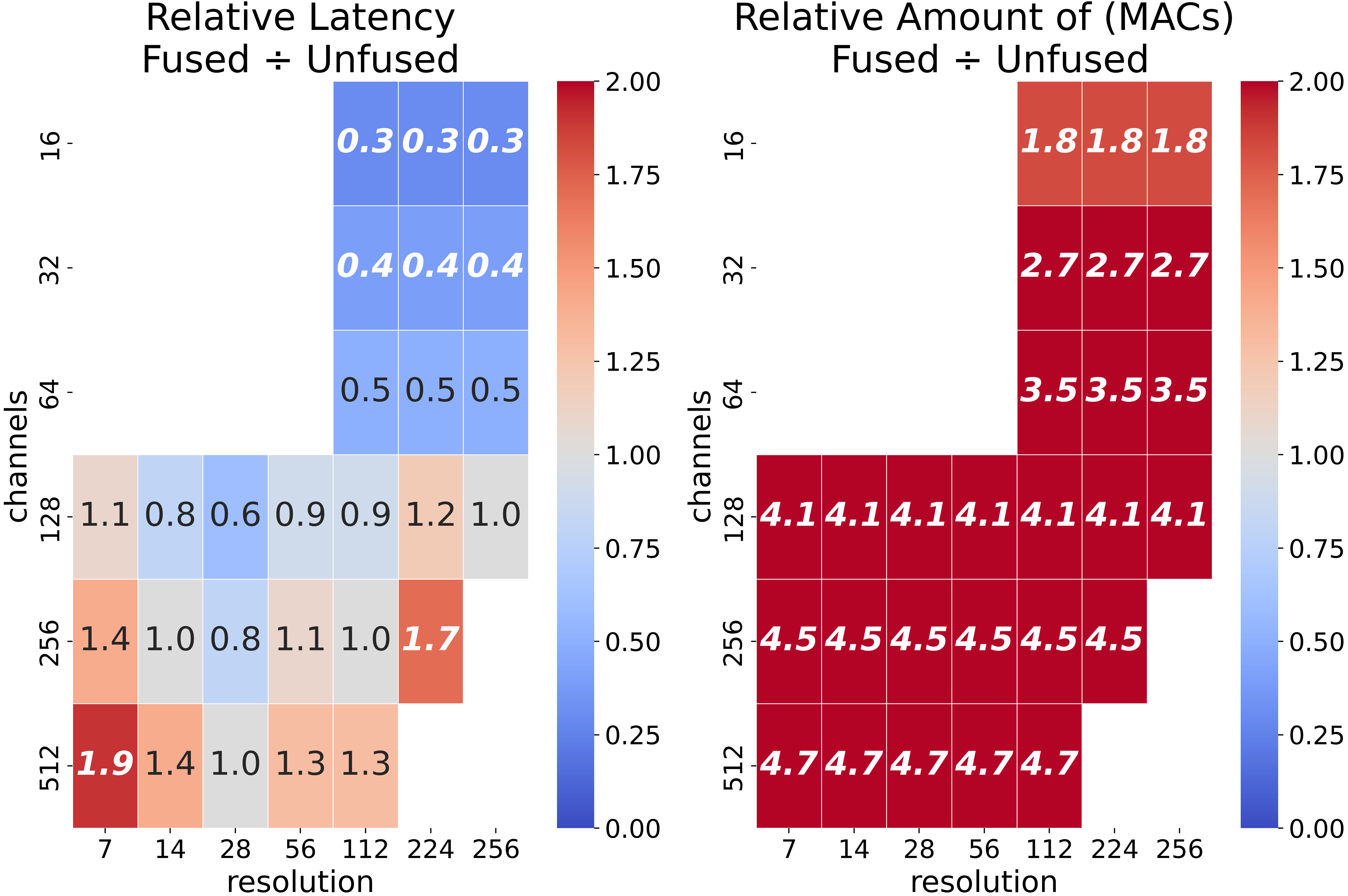}
  \caption{ Latency comparison of fused and unfused MBConv. The missing fields resulted in out-of-memory errors. }
  \label{fig:fusedexplatency}
\end{figure}

In \Cref{fig:fusedexplatency} we measured relative latency of the fused and unfused MBConv \cite{mobilenetv2}. The fused MBConv retains a lower latency for many configurations of channel dimension and operating resolution (value below 1 in \Cref{fig:fusedexplatency}), even though its higher amount of MACs (value over 1 in \Cref{fig:fusedexplatency}).
Mainly for channel dimensions higher than 256, latency increases compared to the unfused MBConv.
Some entries are missing due to OOM (out of memory) errors.

\section{Resolution Scaling}
In \Cref{fig:avgexeresolution} we evaluated execution time of various models with different input resolutions. We used a batch size of 200, the same as for throughput calculation. Even though LowFormer-B1 has a higher or similar top-1 accuracy, it retains a lower average execution time than EfficientMod-xs,\cite{efficientmodulation} FastViT-T8 \cite{fastvit}, RepViT-M1.1 \cite{repvit} and the LowFormer-B1 version without downsampling in the attention block, independent of the input resolution.

{\small
\bibliographystyle{ieee_fullname}
\bibliography{egbib}
}